\documentclass{article}

\usepackage{microtype}
\usepackage{graphicx}
\usepackage{caption}
\usepackage{subcaption}
\usepackage{booktabs} 

\usepackage{hyperref}

\usepackage[accepted]{icml2021}

\icmltitlerunning{Graph Classification by Mixture of Diverse Experts}

\usepackage{mathtools}
\DeclarePairedDelimiterX{\infdivx}[2]{(}{)}{%
  #1\;\delimsize\|\;#2%
}

\usepackage{amsmath,amsthm,verbatim,amssymb,amsfonts,amscd,mathrsfs}
\theoremstyle{plain}
\newtheorem{theorem}{Theorem}

\theoremstyle{remark}

\theoremstyle{definition}

\usepackage[inline]{enumitem}

\begin{document}

\twocolumn[
\icmltitle{Graph Classification by Mixture of Diverse Experts }



\icmlsetsymbol{equal}{*}

\begin{icmlauthorlist}
\icmlauthor{Fenyu Hu}{equal,cri,sch}
\icmlauthor{Liping Wang}{equal,cri,sch}
\icmlauthor{Shu Wu}{cri,sch}
\icmlauthor{Liang Wang}{cri,sch}
\icmlauthor{Tieniu Tan}{cri,sch}
\end{icmlauthorlist}

\icmlaffiliation{cri}{Center for Research on Intelligent Perception and Computing, Institute of Automation, Chinese Academy of Sciences}
\icmlaffiliation{sch}{School of Artificial Intelligence, University of Chinese Academy of Sciences}

\icmlcorrespondingauthor{Shu Wu}{shu.wu@nlpr.ia.ac.cn}

\icmlkeywords{Machine Learning, ICML}

\vskip 0.3in
]



\printAffiliationsAndNotice{\icmlEqualContribution} 
\begin{abstract}
Graph classification is a challenging research problem in many applications across a broad range of domains. In these applications, it is very common that class distribution is imbalanced.
Recently, Graph Neural Network (GNN) models have achieved superior performance on various real-world datasets. Despite their success, most of current GNN models largely overlook the important setting of imbalanced class distribution, which typically results in prediction bias towards majority classes.
To alleviate the prediction bias, we propose to leverage semantic structure of dataset based on the distribution of node embedding. Specifically, we present GraphDIVE, a general framework leveraging mixture of diverse experts (i.e., graph classifiers) for imbalanced graph classification. With a divide-and-conquer principle, GraphDIVE employs a gating network to partition an imbalanced graph dataset into several subsets.
Then each expert network is trained based on its corresponding subset. Experiments on real-world imbalanced graph datasets demonstrate the effectiveness of GraphDIVE. 

\end{abstract}

\section{Introduction}
\label{Introduction}
Graph classification aims to identify class labels of graphs in a dataset, which is a critical and challenging problem for a broad range of real-world applications \cite{huang2016learning,fey2018splinecnn,zhang2019inductive,jia2020graphsleepnet}.  For instance, in chemistry, a molecule could be represented as a graph, where nodes denote atoms, and edges represent chemical bonds. Correspondingly, the classification of molecular graphs can help predict  target molecular properties \cite{ogb_2020_nips}.

\begin{figure}
    \centering
		\begin{subfigure}[t]{.45\linewidth}
			\centering
			\includegraphics[height=2.5cm]{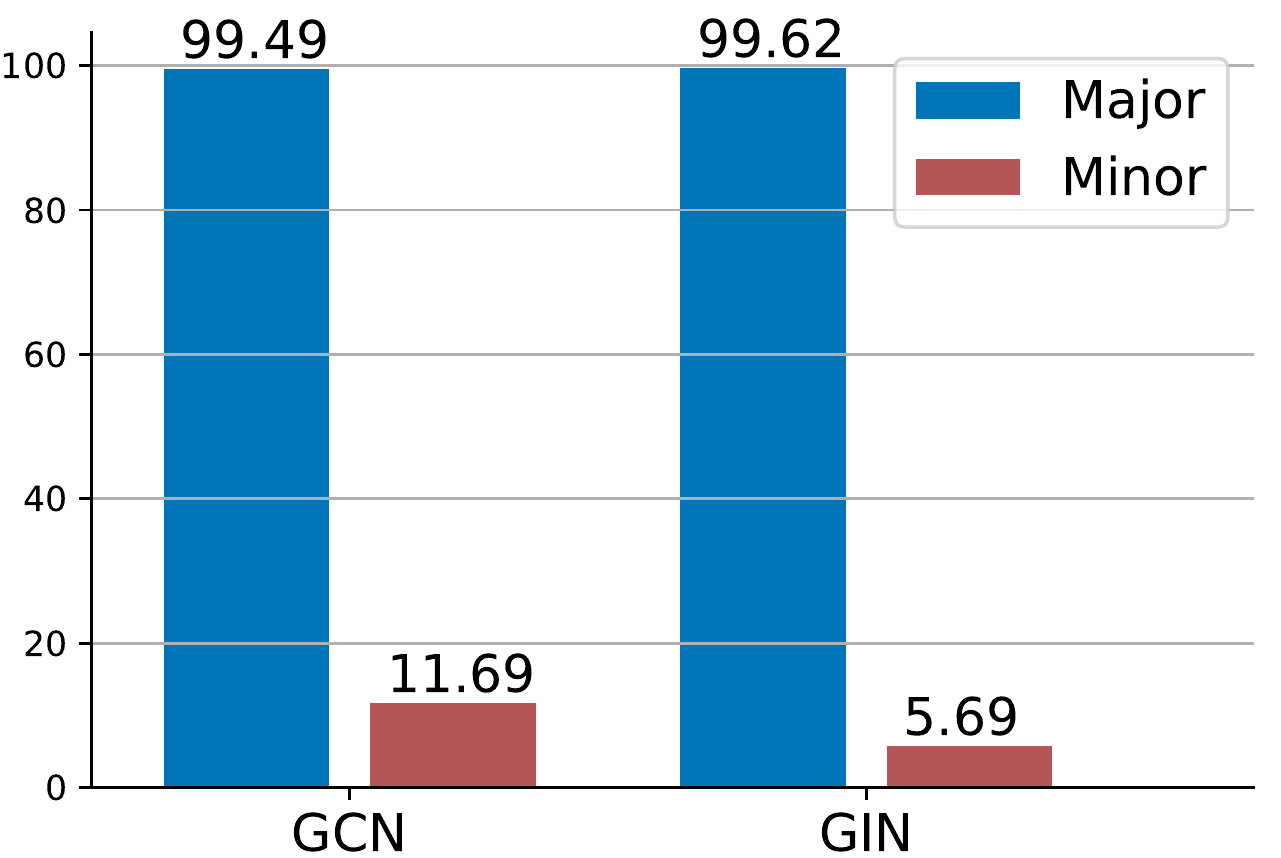} %
			\caption{\small \textbf{ Accuracy}}
	\label{fig:single}
		\end{subfigure}
		\hfill
		\begin{subfigure}[t]{.45\linewidth}
			\centering
			\includegraphics[height=2.5cm]{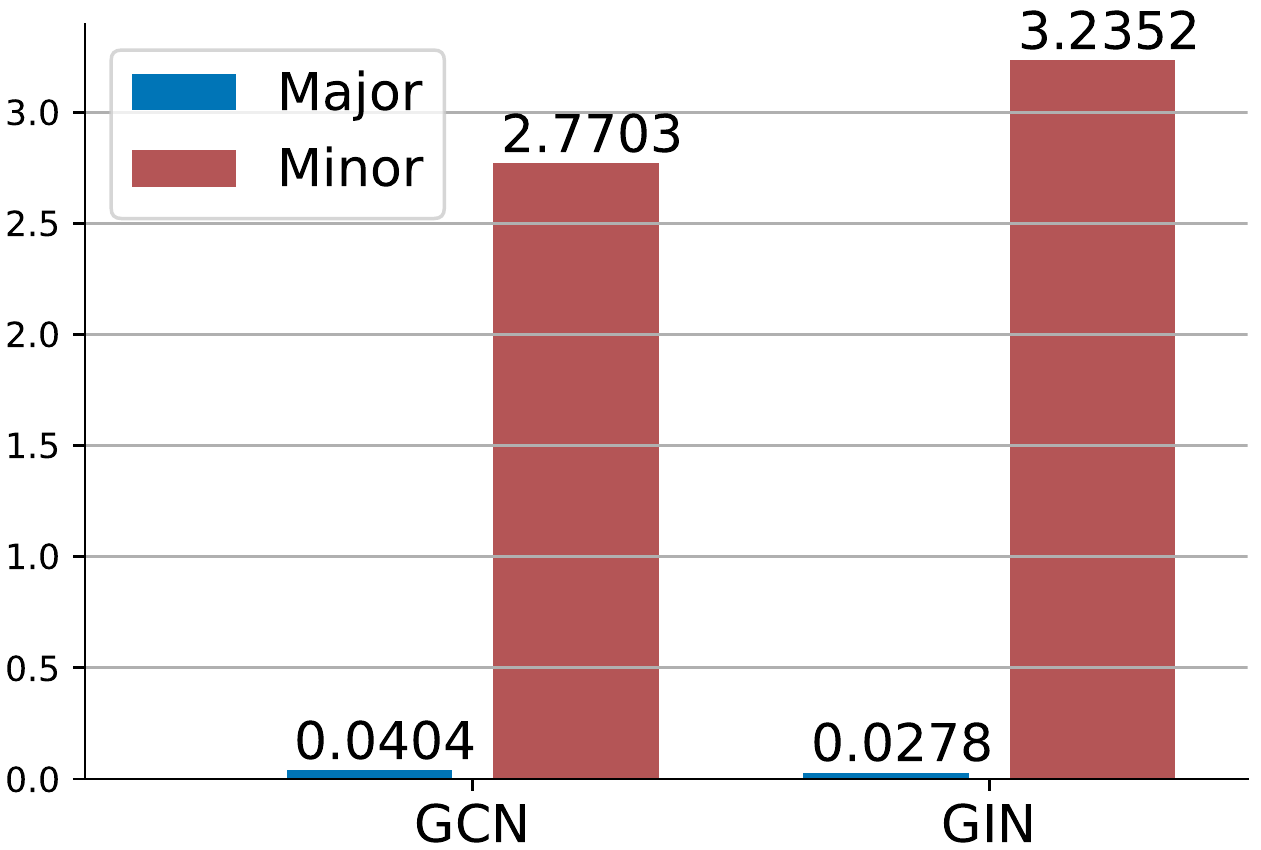} %
			\caption{\small \textbf{ Cross-entropy loss}}
			\label{fig:single}
		\end{subfigure}
		\hfill
\caption{Test accuracy and cross-entropy loss on OGBG-HIV dataset.}
\label{fig:accloss}
\end{figure}

As a powerful approach 
to 
graph representation learning, Graph Neural Network (GNN) models have achieved outstanding performance in graph classification \cite{ying2018hierarchical,xu2018gin,pmlr-v119-wang20m,corso2020principal}. Most of existing GNN models first transform nodes into low-dimensional dense embeddings to learn discriminative graph attributive and structural features, and then summarize all node embeddings to obtain a global representation of the graph. Finally, Multi-Layer Perceptrons (MLPs) are used to facilitate end-to-end training. Nevertheless, current GNN models largely overlook the important setting of imbalanced class distribution, which is ubiquitous in practical graph classification scenarios. For example, in OGBG-MOLHIV dataset \cite{ogb_2020_nips}, only about 3.5\% of molecules can inhibit HIV virus replication. Figure \ref{fig:accloss} presents graph classification results of GCN \cite{gcn} and GIN \cite{xu2018gin} on this dataset. Considering either test accuracy or cross-entropy loss, the classification performance of  minority class falls far behind that of majority class, which indicates the necessity of boosting GNN from the perspective of  imbalanced learning.               

However, apart from suffering the well-known learning bias towards majority classes \cite{sun2009classification,he2009learning,kang2019decoupling}, the class-imbalanced learning problem is exacerbated on graph classification due to the following reasons:
(\textit I) \emph{structure diversity}; (\textit {II}) \emph{poor applicability in multi-task classification setting}. First, structure diversity and the related out-of-distribution problem is very ubiquitous in real-world graph datasets \cite{hu2019strategies}. For example, when scientists want to forecast COVID-19 property, it will be difficult because COVID-19 is different from all known virus. This means typical imbalanced learning methods, such as re-sampling \cite{chawla2002smote} and re-weighting \cite{huang2016learning}, may no longer work on graph datasets. Because of the potential over-fitting 
to minority classes, these imbalanced strategies are sensitive to fluctuations of minor classes, resulting in unstable performance. Second, in more common cases, such as drug discovery
\cite{ogb_2020_nips,becigneul2020optimal,pan2015joint} and functional brain analysis \cite{pan2016task},  graph datasets contain multiple classification tasks due to need  for  predicting  various properties of a graph simultaneously. For example, Tox21 Challenge\footnote{\url{https://tripod.nih.gov/tox21/challenge/}} aims to predict whether certain chemical compounds are active for twelve pathway assays, which can be viewed as twelve-task binary classification setting.
Existing imbalanced learning methods are originally designed for single-task classification. Therefore, it is difficult to apply existing imbalanced learning strategies \cite{chawla2002smote,kang2019decoupling,kim2020m2m} to multi-task setting.

\begin{figure}
\centering
\centerline{\includegraphics[width=\linewidth]{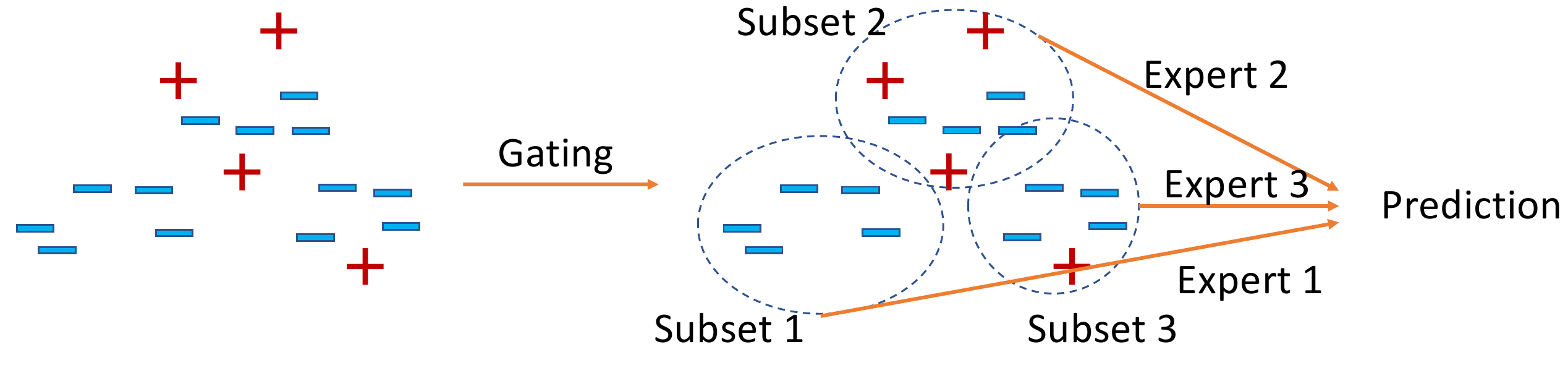}}
\caption{Visual illustration of GraphDIVE, consisting of the gating network and expert network.}
\label{fig:divide}
\end{figure}

To this end, we propose a novel imbalanced \underline{Graph} classification framework with \underline{DIV}erse \underline{E}xperts, referred to as GraphDIVE for brevity.
At first, we leverage a gating network to capture the semantic structure of the dataset. As illustrated in Figure \ref{fig:divide}, the semantically similar graphs are grouped into the same subset.
Then, multiple classifiers, which are referred to as experts, are trained based on their corresponding subsets.
Due to the difference in semantic structure,  samples of majority class and minority class tend to be concentrated in different subsets. Obviously, for the subset containing most of the minority class (please refer to Subset 2 in Figure \ref{fig:divide}), the imbalance phenomenon is alleviated.
As a result, the performance of minority class get promoted.

We systematically study the effect of GraphDIVE  on public benchmarks and obtain  the following key observations: (\textit I) existing imbalanced learning strategies are difficult to offer satisfactory  improvement on graph datasets because of the  graph diversity problem, (\textit {II}) the performance of existing GNNs on graph classification can be further improved by appropriately modeling the  imbalanced distribution, (\textit {III})
capturing semantic structure of datasets can address the structure diversity problem and alleviate the bias towards majority classes, (\textit {IV}) different gates are generally necessary for multi-task setting. Apart from these observations, GraphDIVE achieves state-of-the-art results in both  single-task and multi-task settings. As an instance, on HIV  and BACE benchmarks, GraphDIVE achieves improvements over GCN by  2.09\% and 4.24\%, respectively.

\section{Related Work}
\label{related_work}

\subsection{Graph Neural Networks}
Over the past few years, we have witnessed the fast development of graph neural networks. They process permutation-invariant graphs with variable sizes and learn to extract discriminative features through a recursive process of transforming and aggregating representations from neighbors. At first, GNNs were introduced by \cite{gori2005new} as a form of recurrent neural networks. Then, \citet{spec-not-general} define the convolutional operations using Fourier transformation and Laplacian matrix. GCN \cite{gcn} approximate the Fourier transformation process by truncating the Chebyshev polynomial to 
the first-order neighborhood. GraphSAGE \cite{sage} samples a fixed number of neighbors and employs several aggregation functions. GAT \cite{gat} aggregates information from neighbors by using attention mechanism. DR-GCN \cite{ijcai2020-398} applies  class-conditioned adversarial network to alleviate bias in imbalanced node classification task. Recently, \citet{GCNII2020} try to tackle the over-smoothing problem by using initial residual and identity mapping. 
Apart from innovation on convolution filters, there are also two main branches in the research of graph classification. For one thing, Graph pooling methods, such as DiffPool \cite{ying2018hierarchical}, Graph U-nets \cite{gunet}, and self-attention pooling \cite{lee2019selfatt}, are developed to extract more global information. \citet{mesquita2020rethinking} take a step further in understanding the effect of local pooling methods. For another, there is recently a growing class of graph isomorphic methods \cite{xu2018gin,morris2019weisfeiler,corso2020principal} which aim to quantity representation power of GNNs. 

\subsection{Class-imbalanced Learning}
\textbf{Re-sampling.} Re-sampling methods aim to balance the class distribution by controlling each class's sample frequencies. It can be achieved by over-sampling or under-sampling \cite{chawla2002smote,han2005borderline,he2008adasyn}. Nevertheless, traditional random sampling methods usually cause  over-fitting in minority classes or under-fitting in majority classes. Recently, \citet{kang2019decoupling} propose to use re-sampling strategy in a two-stage training scheme. Besides, \citet{liu2020deep}, \citet{kim2020m2m}, and \citet{chou2020remix} generate augmented samples to supplement minority classes, which can also be viewed as re-sampling methods. However, it is a non-trivial task to apply augmentation on graphs with variable sizes. These re-sampling methods are also unable to produce multiple predictions simultaneously in multi-task setting because different tasks are usually with different class distributions. 

\textbf{Re-weighting.} Re-weighting methods generally assign different weights to different samples. Traditional scheme re-weights classes  proportionally to the inverse of the class frequency, which tends to make  optimization difficult under extremely imbalanced settings \cite{huang2016learning,huang2019deep}. Another line of work assigns weights according to the properties of each training instance. FocalLoss \cite{focalloss} lowers the weights of the well-classified samples. GHM \cite{li2019gradient} improves FocalLoss by further lowering the weights of very large gradients considering outliers. However, these two kinds of methods need  prerequisite of domain experts to hand-craft the loss function in specific task, which may restrict their applicability. Recently, \citet{cui2019class} introduce the effective number of samples to put larger weights on minority classes. \citet{tan2020equalization} propose an equalization loss function that randomly ignores gradients from majority classes.  LDAM \cite{ldam} introduces a label-distribution-aware loss function that encourages larger margins for minority classes. 
Despite the simplicity in implementation, re-weighting methods do not consider the semantic structure of  datasets. So, they may not handle the graph structure diversity problem, causing unstable predictions.

\subsection{Mixture of Experts}

Mixture of Experts (MoE) is mainly based on divide-and-conquer principle, in which the problem space is first divided and then is addressed by specialized experts \cite{expert91}. MoE has been explored by several researchers and has witnessed success in a wide range of applications \cite{tresp2001mixtures,collobert2002parallel,masoudnia2014mixture,eigen2013learning}. In recent years, there is a surge of interest of in incorporating MoE models to address challenging tasks in natural language processing and 
computer vision. \citet{shazeer2016outrageously} introduce a sparsely-gated MoE network, which improves the performance of machine translation and language modeling. \citet{shen2019mixture} evaluate the translation quality and diversity of MoE models. \citet{fedus2021switch} simplify the computational costs of MoE and make it possible to train models with trillion parameters. In computer vision, there are some frameworks \cite{ge2015subset,ge2016fine} which have demonstrated the effectiveness of MoE in fine-grained classification. 
Contrary to existing MoE methods which focus on increasing model capacity, we find MoE are surprisingly overlooked in graph machine learning and that MoE are 
especially appropriate for imbalanced graph classification task. We also propose two variants considering posterior and prior distributions for the gating function. There are also concurrent MoE works \cite{ma2018modeling,qin2020multitask,tang2020progressive} that involve multi-task learning. However, they do not consider the important and ubiquitous class-imbalance problem.

\section{Proposed Method: GaphDIVE}

The technical core of GraphDIVE is the notion to leverage the semantic structure of the graph dataset based on the node embedding distributions. This notion encourages the GNN to group the  structurally different but property-similar graphs into the same subset. Then the minority classes are more likely to be classified correctly by
certain experts. 
Our method is similar to traditional MoE \cite{expert91}, but with a distinct motivation for imbalanced graph classification.  In the following, we first present preliminaries and then describe the algorithmic details of GraphDIVE. Finally, we present two 
model variants for multi-task setting.

\subsection{Preliminaries} 
Let $D=\{(G_1, {Y}_1),\dots, (G_n, {Y}_n)\}$ denote training data, where $G_i=(A_i, X_i, E_i)$ denotes a graph containing adjacency matrix, node attribute matrix and edge attribute matrix respectively. $Y_i=(y_1, \dots, y_T)$ represents the labels of $G_i$ across $T$ tasks. The task of graph classification is to learn a mapping $f:\ G_i \rightarrow {Y_i}$. In this paper, we only consider the binary classification situation that exists widely in practical applications \cite{yanardag2015deep,ogb_2020_nips}. 
For the class-imbalanced problem, the number of instances of majority class is far more than that of minority class.

	\begin{figure*}
		\centering
		\begin{subfigure}[t]{.3\linewidth}
			\centering
			\includegraphics[height=5cm]{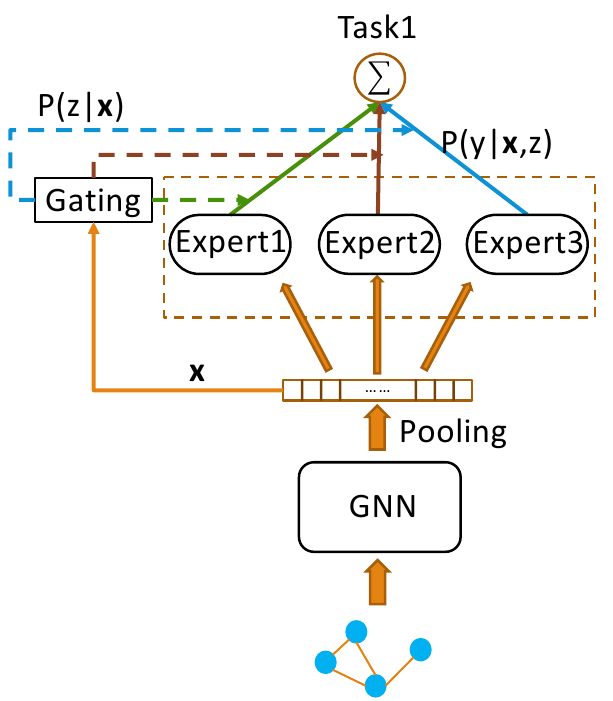} %
			\caption{\small \textbf{ GraphDIVE for single task}}
			\label{fig:single}
		\end{subfigure}
		\hfill
		\begin{subfigure}[t]{.3\linewidth}
			\centering
			\includegraphics[height=5cm]{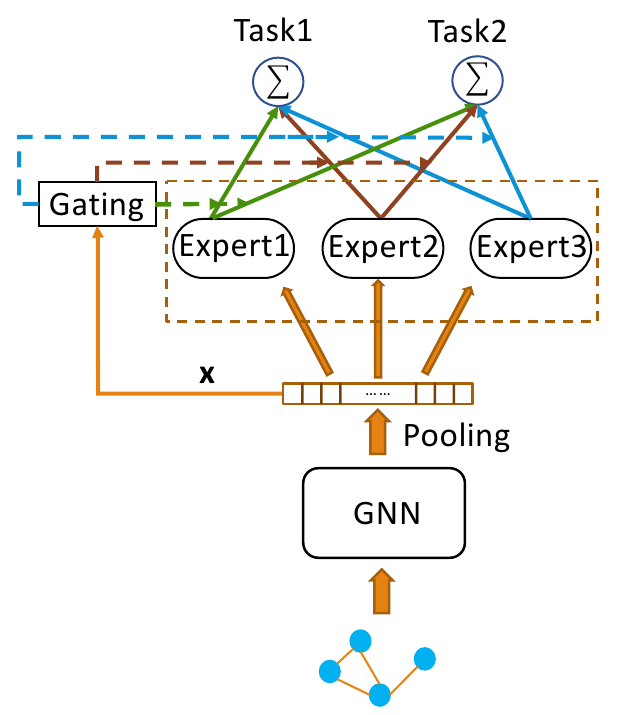} %
			\caption{\small \textbf{   GraphDIVE with shared gates}}
			\label{fig:shared} 
		\end{subfigure}
		\hfill
		\begin{subfigure}[t]{.35\linewidth}
			\centering
			\includegraphics[height=5cm]{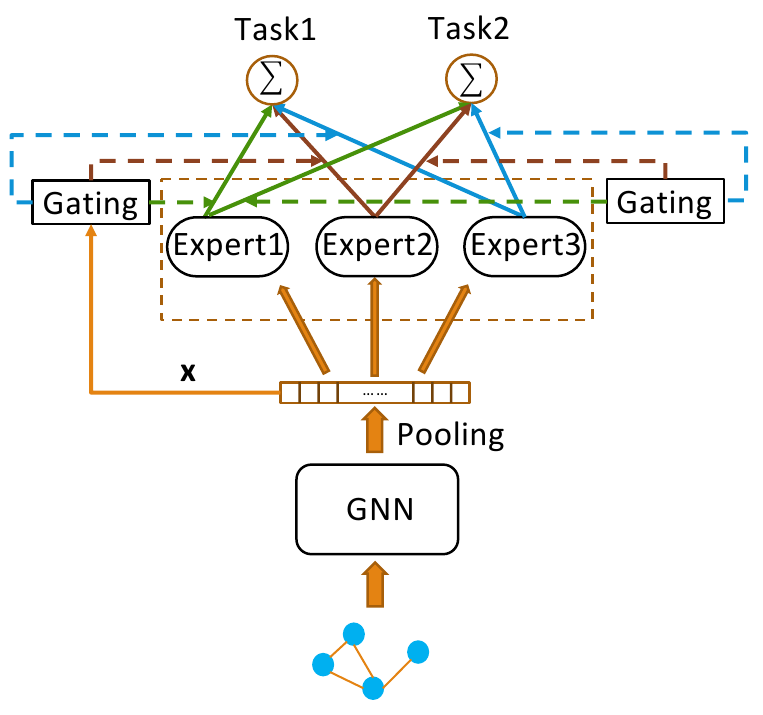} %
			\caption{\small \textbf{GraphDIVE with individual gates}}
			\label{fig:individual}
		\end{subfigure}

		\caption{\label{fig:model}\small
			Model variants in different settings. The latter two variants are designed for multi-task setting. The crossing of the dotted and solid lines of the same color indicates multiplication operation.}
		\vspace*{-4mm}
	\end{figure*}

\subsection{Architecture}
\label{method:arch}
The ubiquitous class-imbalanced problem of graph datasets brings a huge challenge to existing GNNs because the classifier will inevitably produce a biased prediction towards majority classes \cite{sun2009classification,he2009learning,tde_2020_nips}. We conjecture that it might be too difficult for one classifier to discriminate all graphs. Inspired by Mixture of Experts (MoE) \cite{expert91}, we propose to assign different experts to different subsets.
As illustrated in Figure \ref{fig:single}, GraphDIVE consists of the following components.

\textbf{Feature extractor.} Similar to the practice in \citet{ogb_2020_nips}, we design a five-layer graph convolution network to extract graph features. Formally, at the $k$-th layer, the representation of the node $v$ is:
\begin{eqnarray}
  & h_{v}^{(k)}=  \operatorname{COMBINE}^{(k)}\left(h_{v}^{(k-1)}, m_{v}^{(k)}\right), \nonumber \\
  & m_{v}^{(k)}= \text { AGGREGATE }^{(k)}\left(h_{v}^{(k-1)}, h_{u}^{(k-1)}, e_{u v}\right),
\end{eqnarray}
where $u \in \mathcal{N}(v)$ denotes the neighbors of node $v$, $h_{v}^{(k)}$ is the representation of $v$ at the $k$-th layer, $e_{uv}$ is the feature vector of the edge between $u$ and $v$, and $m_{v}^{(k)}$ is the message aggregated to node $v$.
On top of the graph feature extractor, we summarize the global representation of a graph $G$ by using a
graph average pooling layer (i.e., readout function):
\begin{equation}
  \boldsymbol {x} = Pooling \left[GNN(G) \right],
\end{equation}
where $\boldsymbol {x}\in\mathbb{R}^{d}$, and $d$ denotes the hidden dimension of graph embedding.
Notably, GraphDIVE is generic to the choice of underlying GNN. Without loss of generality, in this paper, we choose two commonly used methods GCN \cite{gcn} and GIN \cite{xu2018gin} as feature extractor. 

\textbf{Mixture of diverse experts.}
Under the assumption that one classifier is difficult to learn the desired mapping in skewed distribution, we adopt a gating network to decompose the imbalanced graph dataset into several subsets. Then a diverse set of individual networks, referred to as experts, are trained for discriminating graphs in their corresponding subsets. This divide-and-conquer strategy makes the learning process easier for each expert, and thus alleviates the bias towards majority class. 

Formally, given a graph with global representation  $\boldsymbol x$ and  label $y$. We introduce a latent variable $z\in\{1, 2, \dots, M\}$, where $M$ represents the number of experts. In GraphDIVE, we decompose the likelihood $p(y|\boldsymbol x;\boldsymbol \Theta)$ as 
\begin{equation}\label{eq:margin}
p(y|\boldsymbol x;\boldsymbol \Theta) = \sum_{z=1}^M p(y, z|\boldsymbol x;\boldsymbol \Theta) =  \sum_{z=1}^M p(z|\boldsymbol x;\boldsymbol \Theta)p(y|z, \boldsymbol x;\boldsymbol \Theta),  
\end{equation}
where $\boldsymbol \Theta=\{\boldsymbol {W}_g \in \mathbb{R}^{d\times M}, \boldsymbol {W}_e \in \mathbb{R}^{d\times 2}\}$ denotes learnable parameters of gating network and expert networks. $\sum_{z=1}^M p(z|\boldsymbol x;\boldsymbol \Theta) =1$, and
$p(z|\boldsymbol x;\boldsymbol \Theta)$ is the output of the gating network, indicating the \emph{prior probability} to assign $\boldsymbol x$ to the $z$-th expert. $p(y|z,\boldsymbol x;\boldsymbol \Theta)$ represents output distribution of the $z$-th expert.

For simplicity, we implement each expert with one linear projection layer followed by a sigmoid function:
\begin{equation}\label{eq:expert}
p(y|z, \boldsymbol x;\boldsymbol {\Theta}) = \sigma (\boldsymbol x \ \boldsymbol {W}_e).
\end{equation}

More specifically, the gating network generates an input-dependent soft partition of the dataset based on cosine similarity between graphs and gating prototypes:
\begin{equation}
 p(z|\boldsymbol x;\boldsymbol \Theta) = softmax \Big(
 \frac{\boldsymbol x \ {\boldsymbol {W}_g^z}}{\tau} \Big),
\end{equation}

where $\tau$ is the temperature hyper-parameter tuning the distribution of $z$, and ${\boldsymbol {W}_g^z}$ is the $z$-th gating prototype. 
Since  samples of minority class and majority class are usually different in semantics, they are likely to be grouped into different subsets. 
For the subset containing most of the minority class, the imbalanced problem phenomenon is much alleviated. 

Besides, unlike existing imbalanced learning strategies which suffer fluctuations in graph structure, GraphDIVE can also group structurally-different but semantically-similar graphs into same subsets. In other words, the semantic structure of the dataset is captured by gating network. Hence, the proposed method can alleviate the above-mentioned \emph{structure diversity} problem of  graph datasets.

\textbf{Prior or posterior distribution.} Apart from using prior probability in Eq. (\ref{eq:margin}), we also consider a model variant  using posterior probability as expert weights. According to Bayes' theorem, posterior probability can be calculated as:
\begin{equation}\label{eq:posterior}
    p(z|\boldsymbol x, y;\boldsymbol \Theta) = \frac{p(z|\boldsymbol x;\boldsymbol \Theta)p(y|z,\boldsymbol x;\boldsymbol \Theta)}{\sum_{z'}p(z'|\boldsymbol x;\boldsymbol \Theta)p(y|z',\boldsymbol x;\boldsymbol \Theta)}.
\end{equation}
As opposed to prior distribution which considers only graph features, this Bayesian extension considers the information from both graph labels and  experts. 
For the convenience of expression, We refer to these two model variants as GraphDIVE-pri and GraphDIVE-post.  



\subsection{Optimization}
\label{subsec:optimization}
We first present the optimization regime of  Bayesian variant. Since the training objective is to maximize the log-likelihood, the loss function can be formulated as following:
\begin{align}\label{eq:exp}
   \mathcal{L}_{post}= -\sum_{z=1}^M & p(z|\boldsymbol x, y;\boldsymbol \Theta) \log p(y|\boldsymbol x,z;\boldsymbol \Theta) 
\end{align}

Noticing the interdependence between posterior distribution and  prediction distribution from experts, we propose to use EM algorithm \cite{em1977} to optimize Eq. (\ref{eq:posterior}) and Eq. (\ref{eq:exp}) iteratively:

\textbf{E-step:} estimate the weight $p(z|\boldsymbol x, y)$ for each expert according Eq. (\ref{eq:posterior}) under current value of parameters $\boldsymbol \Theta$.

\textbf{M-step:} update  parameters $\boldsymbol \Theta$ using stochastic gradient descent algorithm, and the gradient $\nabla \log p(y, z|\boldsymbol x)$ is weighted by the estimated $p(z|\boldsymbol x, y)$. The gradients are blocked in the computation of posterior distribution.

What is more, similar to the practice in \citet{hasanzadeh2020bayesian}, we introduce a Kullback--Leibler (KL) divergence regularization term  $\mathrm{KL}(p(z|\boldsymbol x, y)||p(z|\boldsymbol x;\boldsymbol \Theta))$ into the final loss function:
\begin{align}\label{eq:loss}
\begin{split}
   \mathcal{L}_{post}= -\sum_{z=1}^M & p(z|\boldsymbol x, y;\boldsymbol \Theta) \log p(y|\boldsymbol x,z; \boldsymbol \Theta) \\
    &+ \lambda * \mathrm{KL}(p(z|\boldsymbol x, y;\boldsymbol \Theta)||p(z|\boldsymbol x;\boldsymbol \Theta)),
\end{split}
\end{align}
where $\lambda$ is a hyper-parameter which controls the extent of regularization. The above KL term ensures the posterior distribution does not deviate too far from the prior distribution. 
So we choose to compute gradients in M-step based on Eq. (\ref{eq:loss}). 




For the optimization of GraphDIVE-pri, the posterior distribution $p(z|x_i, y_i;\boldsymbol \Theta)$ in Eq. (\ref{eq:loss}) is replaced with prior distribution $p(z|x_i;\boldsymbol \Theta)$. And regularization item becomes zero. In this case, the gating network and the expert network can be jointly optimized according to the following  objective:
\begin{equation}
    \mathcal{L}_{pri}= -\sum_{z=1}^M  p(z|\boldsymbol x;\boldsymbol \Theta) \log p(y|\boldsymbol x, z; \boldsymbol \Theta).
\end{equation}
In experiment, we find both variants fit graph data well and demonstrate superior generalization ability. And detailed comparison can be found in Section \ref{exp:comp}. 
\subsection{Adaptation to multi-task scenario}
\label{method:multi}
In real-life applications, researchers are likely to be confronted with imbalanced graph classification in multi-task setting. 
Multi-task setting increases the difficulty of imbalanced learning as one graph might be of the majority class for some tasks while being of  minority class for the other tasks.
Existing imbalanced learning methods are originally designed for single-task classification, so it is a non-trivial task to adapt them for multi-task setting. In this paper, we consider two options for multi-task scenario, which are illustrated in Figure \ref{fig:shared} and Figure \ref{fig:individual}. 

\textit{Option I: Shared gates.} This is the simplest adaptation variant, which uses shared gating weights for different tasks (see Figure \ref{fig:shared}). Compared with the model in Figure \ref{fig:single}, the only difference is that each expert generates predictions for different tasks. This variant has the  advantage that it reduces model parameters, especially in settings with many experts. However, empirically this variant does not work well in our setting. Our reasoning is that it implicitly assumes training instances obey the same label distributions across all tasks.

\textit{Option II: Individual gates.} Another natural option is to assign  different groups of gating weights for different tasks, as illustrated in \ref{fig:individual}. Compared to the \emph{shared gates} solution, this variant has several additional gating networks. Notably, it models  task relationships in a sophisticated way. For two less related tasks, the sharing expert weights will be penalized, resulting in different expert weights instead.

\section{Theoretical Analysis}
In this section, theoretical analysis is provided from variational inference perspective to help understand why GraphDive works. 

Assuming that an observed graph $\boldsymbol x$ is related to a latent variable $z=\{1, 2, \dots, M\}$, and $p(z|\boldsymbol x)$ denotes the probability of $\boldsymbol x$ locating in the $z$-th sub-region of  feature space. Let $q(z|\boldsymbol x, y)$ denote a variational distribution to infer latent variable given  observed data, and $p(y|\boldsymbol x, z;\boldsymbol \Theta)$ denotes the distribution of the prediction of each expert. As for the KL regularization term, we consider the simple case of $\lambda=1$. Then we prove the following theorem.
{\theorem	\label{main_theorem}
In GraphDIVE, optimizing the final loss is equivalent to the optimization of the lower bound of $\log p(y|\boldsymbol x)$:
\begin{align}\label{main_eq}
\begin{split}
\log p(y|\boldsymbol x;\boldsymbol\Theta)& \ge \mathbb{E}_{q(z|\boldsymbol x, y)}\log p(y|\boldsymbol x, z;\boldsymbol\Theta) \\
&- \mathrm {KL}(q(z|\boldsymbol x, y)||p(z|\boldsymbol x;\boldsymbol\Theta))
\end{split}
\end{align}}
{\proof 
\begin{align*}
\begin{split}
&\quad\ \mathrm {KL}(q(z|\boldsymbol x, y)||p(z|\boldsymbol x, y;\boldsymbol\Theta))\\
&=\mathbb{E}_{q(z|\boldsymbol x, y)}\log \frac{q(z|\boldsymbol x, y)}{p(z|\boldsymbol x, y;\boldsymbol\Theta)}\\
&=\mathbb{E}_{q(z|\boldsymbol x, y)}\log \frac{q(z|\boldsymbol x, y)p(y|\boldsymbol x;\boldsymbol\Theta)}{p(z, y|\boldsymbol x;\boldsymbol \Theta)}\\
    &=\log p(y|\boldsymbol x;\boldsymbol\Theta) + \mathbb{E}_{q(z|\boldsymbol x, y)}\log \frac{q(z|\boldsymbol x, y)}{p(y|\boldsymbol x, z;\boldsymbol\Theta)p(z|\boldsymbol x;\boldsymbol\Theta)}\\
    &=\log p(y|\boldsymbol x) - \mathbb{E}_{q(z|\boldsymbol x, y)}\log p(y|\boldsymbol x, z;\boldsymbol\Theta) \\ 
    &\quad+ \mathrm {KL}(q(z|\boldsymbol x, y)||p(z|\boldsymbol x;\boldsymbol\Theta)
\end{split}
\end{align*}
Notice that $\mathrm {KL}(q(z|\boldsymbol x, y)||p(z|\boldsymbol x, y;\boldsymbol\Theta)) \ge 0$, which concludes the proof.
}

Eq. (\ref{main_eq}) provides a lower bound of $\log p(y|\boldsymbol x)$. GraphDIVE optimizes the evidence lower bound (ELBO) from the perspective of variational inference. The more closer is
$p(z|x, y;\Theta)$ to $q(z|x, y)$, the tighter the lower bound is.
The first item on the right hand encourages $q(z|x, y)$ to be high for experts which make good predictions. And the second item is the Kullback-Leibler divergence between the variational distribution and the prior distribution output by the gating network. With this term, the gating network considers both  graph labels and experts' capacity when partitioning  graph datasets.


\section{Experiments}

In this section, we provide extensive experimental results of GraphDIVE on imbalanced graph classification datasets under both single-task and multi-task settings. The experimental results demonstrate  superior performance of GraphDIVE over state-of-the-art models. Besides, we present a case study to demonstrate how the gating mechanism improve classification performance of minority class.

\subsection{Datasets}
We conduct experiments on the  recently released large-scale datasets of \textit{Open Graph Benchmark\footnote{\url{https://ogb.stanford.edu/}}}(OGB) \cite{ogb_2020_nips}, which are more realistic and challenging than traditional graph datasets. More specifically, we choose six molecular graph datasets from OGB: BACE, BBBP, HIV, SIDER, CLIONTOX, and TOX21. These datasets cover different  complex chemical properties, such as inhibition of human $\beta$-secretase, and blood-brain barrier penetration. All these datasets contains two classes for each task. Here we  give a brief 
statistics of these datasets in Table \ref{tbl:dataset}, and a more detailed description can be found in \citet{ogb_2020_nips}. For the multi-class classification task, please refer to supplementary materials for details.

\begin{table}[]
\centering
\caption{Statistics of OGB Datasets}
\label{tbl:dataset}
\small
\resizebox{\linewidth}{!}{
\begin{tabular}{lc|c|c|c|c}\toprule
&Dataset        & Size & \# tasks & Avg. Size & Positive Ratio\\ \hline
&BACE           & 1513 & 1        & 34.1      &      45.6\%         \\
&BBBP           & 2039 & 1        & 24.1      &       23.5\%        \\
&HIV          &41127  & 1        & 25.5       &       3.5\%       \\
&SIDER-task-3   &1427  & 1        & 33.6      &        1.5\%        \\
&CLINTOX        & 1477 & 2        & 26.2      &       6.97\%       \\
&SIDER          & 1427 & 27       & 33.6      &     25.14\% \\ 
&Tox21         & 7831 & 12       & 18.6      &       7.52\%  \\
\bottomrule
\end{tabular}
}
\end{table}

\begin{table}[]
\centering
\caption{Summary of classification ROC-AUC (\%) results under single task setting.}
\resizebox{\linewidth}{!}{
\begin{tabular}{lc|cccc}
\toprule
        & & BACE             & BBBP             & HIV            & SIDER-3                                  \\
\midrule
&GCN      & 79.15$\pm$1.44   & 68.87$\pm$1.51   & 76.06$\pm$0.97 & 36.11$\pm$11.40                          \\
&GIN      & 72.97$\pm$4.00   & 68.17$\pm$1.48   & 75.58$\pm$1.4  & 30.90$\pm$7.68                           \\
&GCN+FLAG & 80.53$\pm$1.43   & 70.04$\pm$0.82   & 76.83$\pm$1.02 & 46.90$\pm$7.22                           \\
&GIN+FLAG & 80.02$\pm$1.68   & 68.60$\pm$1.27   & 76.54$\pm$1.14 & 42.09$\pm$9.76                           \\
&GSN      & 76.53$\pm$4.54   & 67.90$\pm $1.86  & 77.99$\pm$1.00 & 37.26$\pm$5.14                           \\
&WEGL     & 78.06 $\pm$ 0.91 & 68.27 $\pm$ 0.99 & 77.57$\pm$1.11 & 48.88 $\pm$ 12.69\\
\midrule
&GraphDIVE-post & 81.10$\pm$1.86 & 69.65$\pm$0.94&76.51$\pm$1.01 &48.29$\pm$5.99  \\
&GraphDIVE-pri & \textbf{83.39$\pm$0.8}   & \textbf{70.33$\pm$1.24}   & \textbf{78.15$\pm$1.28} & \textbf{48.93$\pm$3.79} \\
\bottomrule
\end{tabular}
}
\label{tbl:gnn}
\end{table}

\begin{table*}[]
\caption{Comparison between GraphDIVE and other imbalanced-learning methods.  The ROC-AUC (\%) values are reported.}
\label{tbl:main_results}
\centering
\resizebox{\linewidth}{!}{
\begin{tabular}{lccccc|cccc}
\toprule
    &~       & \multicolumn{4}{c|}{GCN}                                                                   & \multicolumn{4}{c}{GIN}                                                                   \\ \cline{3-10}
        &   & \textbf{BACE}        & \textbf{BBBP}        & \textbf{HIV}         & \textbf{SIDER-3}     & \textbf{BACE}        & \textbf{BBBP}        & \textbf{HIV}         & \textbf{SIDER-3}     \\ \cline{3-10}
&  & 79.15$\pm$1.44    &  68.87$\pm$1.51 & 76.06$\pm$0.97  & 36.11$\pm$11.40    &   72.97$\pm$4.00  & 68.17$\pm$1.48  &75.58$\pm$1.40  & 30.90$\pm$7.68 \\
\midrule
&Focal Loss &   81.08$\pm$2.02   & 67.90$\pm$1.16                      &   76.56$\pm$1.15 &   29.36$\pm$10.30  &    72.36$\pm$4.59                  &    66.10$\pm$1.65   &   77.00$\pm$1.10 &    26.90$\pm$9.03     \\ 
&GHM        &     80.51$\pm$1.54   &        67.04$\pm$1.26      &  75.33$\pm$1.44                     &  19.90$\pm$6.99       &                70.93$\pm$4.74       &   67.71$\pm$2.26                &      74.21$\pm$1.07            &                 38.57$\pm$9.60                            \\ 
&LDAM       &     78.91$\pm$2.10 & 67.08$\pm$0.94   &       76.58$\pm$1.69  &    38.50$\pm$9.99 &  75.35$\pm$2.94  &  65.89$\pm$2.05  &     74.57$\pm$3.46             &     37.76$\pm$8.46      \\ 
&Decoupling &   80.01$\pm$1.01 &68.42$\pm$1.46  & 76.43$\pm$1.39 & 34.24$\pm$17.02                   &  73.72$\pm$4.55  &67.40$\pm$1.85  & 75.69$\pm$1.43 &   35.07$\pm$12.19    \\
\midrule 
&GraphDIVE  & \textbf{83.39$\pm$0.80}  & \textbf{70.33$\pm$1.24}   & \textbf{78.15$\pm$1.28}   &  \textbf{48.93$\pm3.79$}   & \textbf{78.41$\pm$1.75}   &    \textbf{68.77}$\pm$1.24               &  \textbf{77.64$\pm$1.01}  &       \textbf{42.60$\pm$5.02}                 \\ 
\bottomrule
\end{tabular}}
\end{table*}

Each graph in molecular graph dataset represents a molecule, where nodes are atoms, and edges are chemical bonds. Each node contains a 9-dimensional attribute vector, including atomic number and chirality, as well as other additional atom features such as formal charge and whether the atom is in the ring. Moreover, each edge contains  a 3-dimensional attribute vector, including bond type, bond stereochemistry, and an additional bond feature indicating whether the bond is conjugated.

\subsection{Experimental Setup}
For a fair comparison, we implement our method and all baselines in the same experimental settings as  \citet{ogb_2020_nips}. For both single-task and multi-task datasets, we follow the original scaffold train-validation-test split with the ratio of 80/10/10. The scaffold splitting separates structurally different molecules into different subsets, which provides a more realistic estimate of the model performance in experimental settings \cite{wu2018moleculenet}. We run ten times for each experiment  with random seed ranging from 0 to 9, and report the mean and standard deviation of test ROC-AUC for all methods.

For hyper-parameter setting, we set the embedding dimension to 300, number of layers to 5, and employ the same GNN backbone network structure. We train the model using Adam optimizer \cite{adam14} with initial learning rate of 0.001. For HIV and TOX21 datasets, we train the network for 120 epochs in light of the scale of the dataset. Moreover, for all the other datasets, we train the model for 100 epochs. 
According to the average performance on the validation dataset, we use grid-search to find the optimal value for $M$ (i.e., the number of experts) and $\lambda$. We set the hyper-parameter space of $M$ as [2, 3, 4, 5, 6, 7, 8] and the hyper-parameter space of  $\lambda$ as [0.001, 0.01, 0.1, 1., 10], respectively.  

We implement all our models based on PyTorch Geometric \cite{pyg2019} and run all our experiments on a single NVIDIA GeForce RTX 2080 Ti 12GB. 

\subsection{Single-task Graph Classification }
\label{exp:comp}

\textbf{Setting and baselines.} For single task, we choose BACE, BBBP and HIV datasets, which are initially single-task binary classification. Besides, at random, we pick the third task of SIDER dataset. 

We consider two representative and competitive graph neural networks as feature extractors: GCN \cite{gcn} and GIN \cite{xu2018gin}. To verify the effectiveness of our method, we also compare the following strong and competitive methods: FLAG \cite{kong2020flag}, GSN \cite{gsn2020}, WEGL \cite{wegl2021}. For all these methods, we use official implementation and follow the original setting.

In addition, we compare our method with state-of-the-art methods that are designed for imbalanced or long-tailed problems, including FocalLoss \cite{focalloss}, LDAM \cite{ldam}, GHM \cite{li2019gradient}, and Decoupling \cite{kang2019decoupling}.
For FocalLoss, we follow the original parameter setting. For LDAM and Decoupling, official implementation is adopted, and we carefully tune the hyper-parameters since there is considerable difference between the graph classification datasets and image classification datasets that these methods originally designed for. 
What is more, for GHM method, we set the number of bins to 30, and momentum $\eta$ as $0.9$.

\textbf{Comparison with SOTA GNNs.} We report the ROC-AUC score of state-of-the-art GNN models in Table \ref{tbl:gnn}.
Overall, we can see that the proposed GraphDIVE shows strong performance across all four datasets. GraphDIVE consistently outperforms other GNN models. Particularly, GraphDIVE achieves up to $12.82 \%$ absolute improvement over GCN on SIDER-3 dataset. The strong performance verifies the effectiveness of the proposed mixture of experts framework. 

Besides, comparing GraphDIVE-post and GraphDIVE-pri, we observe that GraphDIVE-pri performs better. We suppose the reason is that the calculation of posterior distribution introduces bias. As formulated in Eq. (\ref{eq:posterior}), posterior distribution considers the capacity of each expert and graph labels. However, the imbalanced distribution of graph labels makes each expert focus more on majority class, hindering the performance of GraphDIVE. On the contrary, Graph-pri reliefs the confounder bias from labels and achieves relatively better results.
In the following text, without specification, we refer to GraphDIVE-pri as GraphDIVE for simplicity.

\textbf{Comparison with SOTA imbalanced learning methods.} We also compare GraphDIVE and other state-of-the-art class-imbalanced learning methods. The results are shown in Table \ref{tbl:main_results}.
Firstly, we find that GraphDIVE  outperforms baseline models (i.e., GCN and GIN) consistently by considerable margins, which implies that the performance of existing GNNs on graph classification can be further improved by appropriately modeling the imbalanced distribution.

It is also worth noting that the state-of-the-art imbalanced learning methods, such as LDAM and Decoupling, do not seem to offer significant or stable improvements over baseline models. For example, Focal loss performs better than GCN on BACE and HIV dataset, but it is inferior to baseline on BBBP and SIDER-3 dataset. We suppose the reason is that either re-sampling or re-weighting methods make the model focus more on minority class, resulting in potential over-fitting  to minority class. When there are distinct differences in test graphs and train graphs, existing imbalanced learning methods may fail to generate accurate predictions. Compared with existing imbalanced learning methods, GraphDIVE usually achieves a higher ROC-AUC score and lower standard deviation. In other words, GraphDIVE maintains remarkable and  stable improvements across different datasets. We attribute this property to the gating and expert networks, which capture semantic structure of the dataset and possess the model superior generalization ability. We also present a case study in Section \ref{exp:experts} to illustrate the effectiveness of GraphDIVE.



\begin{table}[]
\caption{
Summary of classification ROC-AUC (\%) results  under multi-task setting.
}
\label{tbl:multask_results}
\resizebox{\linewidth}{!}{
\begin{tabular}{lc|lll}
\toprule
        & & CLINTOX                               & SIDER            & TOX21           \\
\midrule
&GCN      & 91.73$\pm$1.73                        & 59.60$\pm$1.77 & 75.29$\pm$0.69     \\
&GCN-DIVE-SG &  90.47$\pm$1.38 & 59.48$\pm$0.96   & 74.17$\pm$0.40 \\

&GCN-DIVE-IG &  \textbf{91.98$\pm$1.32} & \textbf{59.62$\pm$0.90}   & \textbf{75.81$\pm$0.65} \\
\midrule
&GIN      & 88.14$\pm$2.51                      & 57.60$\pm$1.40 & 74.91$\pm$0.51     \\
&GIN-DIVE-SG &  82.02$\pm$6.75 & 57.57$\pm$1.25   & 72.78$\pm$0.60 \\
&GIN-DIVE-IG & \textbf{89.52$\pm$1.08}  &\textbf{58.78$\pm$1.28} & \textbf{74.96$\pm$0.60}\\
\bottomrule
\end{tabular}}
\end{table}

\subsection{Multi-task Graph Classification}
For multi-task, we choose three datasets, including TOX21, CLINTOX and SIDER.
Since there is no prior imbalanced learning research on multi-task graph classification and existing imbalanced learning strategies are difficult to be adapted to multi-task setting, we only compare GraphDIVE and baseline models. The results in Table \ref{tbl:multask_results} show GraphDIVE still advance the performance over GCN and GIN. Notably, the improvements under multi-task setting are not as remarkable as those under single-task setting. This observation is expected because different tasks may require the shared graph representation to optimize towards different  directions. 
So the  multi-task imbalanced graph 
classification is a challenging research direction. We also note that the individual-gate variant (i.e., DIVE-IG) generally performs better than shared-gate variant (i.e., DIVE-SG).
Considering that one graph might be of the majority class for some tasks while being of  minority class for the other tasks, this result
indicates that different gates are necessary for multi-task setting.

\subsection{Effectiveness Analysis for Diverse Experts}
\label{exp:experts}

\begin{figure}
\centering
\centerline{\includegraphics[width=\linewidth]{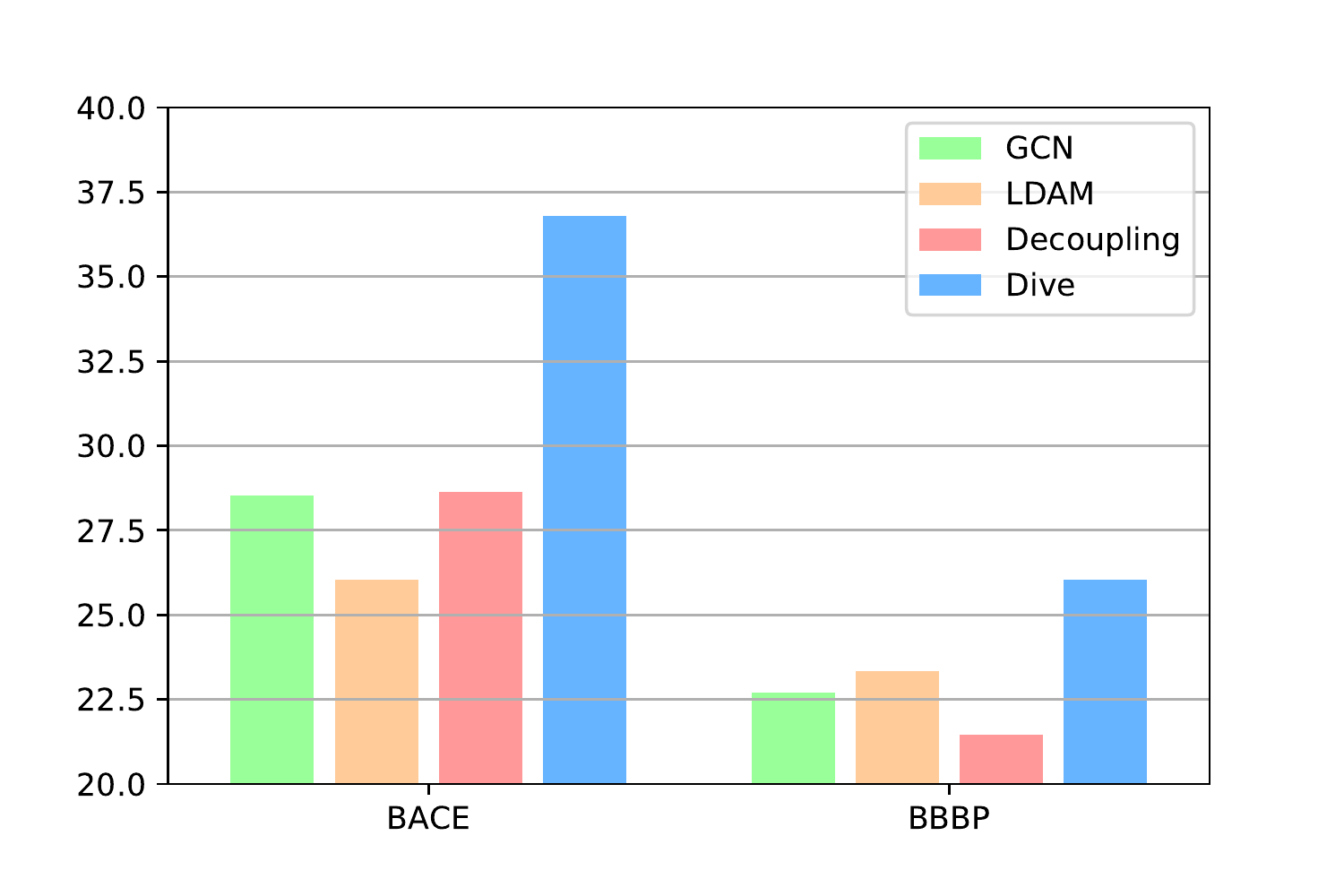}}
\caption{Classification accuracy of minor class on BACE and BBBP dataset.}
\label{fig:minority}
\end{figure}

\begin{figure}
\centering
\centerline{\includegraphics[width=\linewidth]{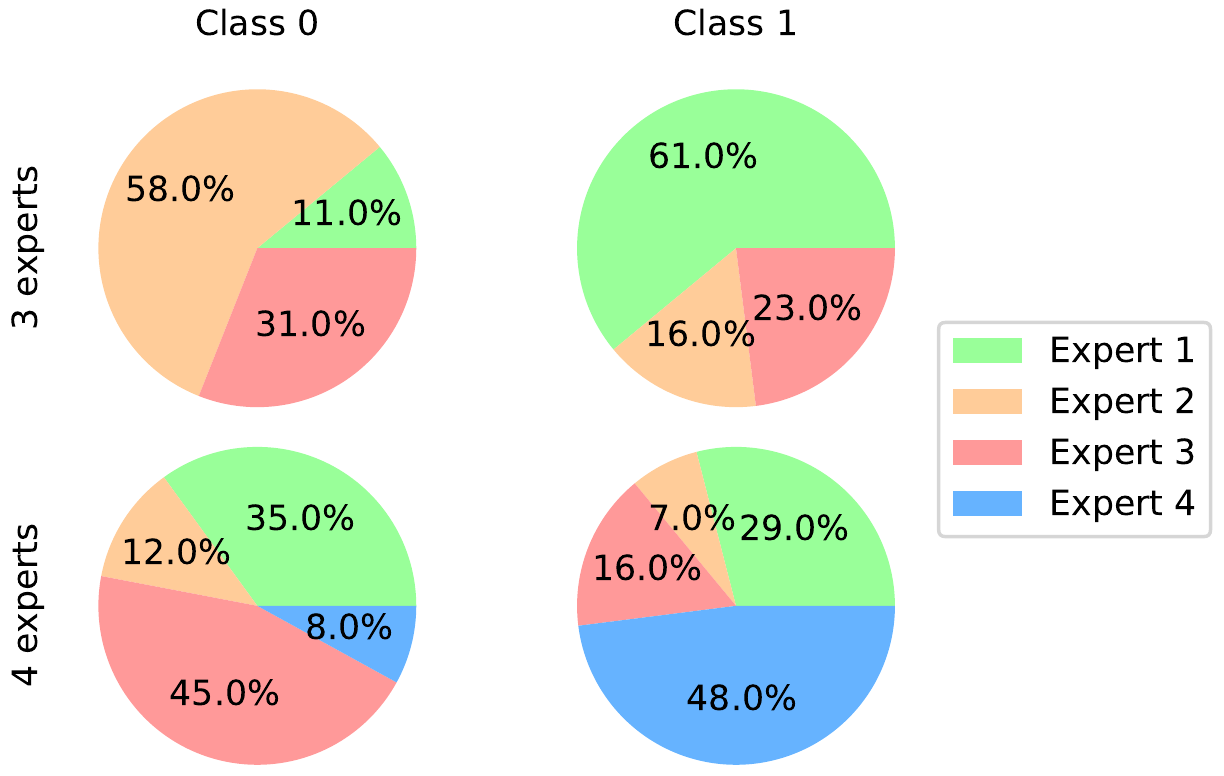}}
\caption{The weights of experts output by gating network on BACE dataset.
The classification of different classes relies on different experts.}
\label{fig:pie_chart}
\end{figure}

To evaluate the effectiveness of the diverse experts in GraphDIVE, we study whether and why diverse experts can alleviate prediction bias towards majority class. More specifically, we report the classification accuracy of minority class in Figure \ref{fig:minority}. Besides, we visualize different experts' predictions on BACE dataset under the setting with three and four experts, respectively.  According to the number of samples assigned to each expert, we present a pie chart in Figure \ref{fig:pie_chart}.


We have the following observations. First of all, existing state-of-the-art re-weighting and re-sampling methods, such as LDAM and Decoupling, have marginal improvements in minority class. This is as expected since they
are prone to over-fitting to minority class and cannot perfectly solve the structure diversity problem.
Secondly, we observe that the proposed GraphDIVE outperforms baseline and existing imbalanced learning methods by a remarkable margin. These improvements suggest that GraphDIVE can successfully alleviate the prediction bias towards majority class and boost performance for minority class. 

Moreover, we take a further step to analyze why GraphDIVE can alleviate the bias. As shown in Figure \ref{fig:pie_chart}, Class 0 and Class 1 denote the majority class and minority class, respectively. It can be observed that the classification of different classes relies on different experts. For example, for the setting with three experts, Expert 1 dominates the classification for the minority class 
while Expert 2 dominates the classification for the majority class. This phenomenon demonstrates that the proposed GraphDIVE successfully captured the semantic structure, i.e., each expert is responsible for a subset of graphs. For the expert which dominates the classification of minority class, its corresponding subset contains more minority class. So training procedure of this 
expert is less likely to be biased towards majority class. 

We also investigate the impact of expert numbers. Please refer to supplementary materials for more detailed information.

\subsection{Effect of gating mechanism}
\begin{table}[]
\caption{Ablation study on gating network.}
\label{tbl:gate_ablation}
\resizebox{\linewidth}{!}{
\begin{tabular}{lc|ccc}
\toprule
         && BACE            & BBBP            & SIDER-3       \\
\midrule
&GCN& 79.15$\pm$1.44    &  68.87$\pm$1.51  & 36.11$\pm$11.40  \\
\midrule
&DIVE-no-gate & 80.99$\pm$0.92 & 68.12$\pm$0.59 & 41.71$\pm$7.50 \\
&GraphDIVE     & \textbf{82.41$\pm$1.37} & \textbf{69.39$\pm$1.21} & \textbf{49.62$\pm$4.44}\\
\bottomrule
\end{tabular}
}
\end{table}

To verify the effectiveness of our proposed gating mechanism, we conduct an ablation experiment
on three single-task datasets. We select GCN as a feature extractor and assign four experts. We replace the gating network with simple arithmetic mean operation on predictions of multiple experts. The model is called Dive-mean accordingly. As Table \ref{tbl:gate_ablation} shows, compared with GraphDIVE, Dive-mean (i.e., DIVE-no-gate) reduces test ROC-AUC by relative 61.33\% on average. Dive-mean may even degrade the performance on BBBP dataset. These results verify the effectiveness of the gating network. Notably, 
DIVE-mean almost shares the same number of parameters as GraphDIVE. Comparing GCN and DIVE-mean, it can also be seen that simply adding network parameters by introducing more experts cannot bring enough improvement. 

To sum up, the above results suggest that both the expert network (refer to Section \ref{exp:experts}) and the gating network are needed to boost the performance of imbalanced graph classification.

\section{Conclusion}
In this paper, we have introduced GraphDIVE, a mixture of  diverse experts framework for  imbalanced graph classification. 
GraphDIVE employs a gating network to learn a soft partition of the dataset, in which process the semantically different graphs are grouped into different subsets. Then diverse experts are trained based on their corresponding subsets. Considering whether to use prior distribution or posterior distribution, we design two model variants and investigate their effectiveness. Besides, we also extend another two variants specially for multi-task setting.
The theoretical analysis shows that GraphDIVE can optimize the exact evidence lower bound with the above divide-and-conquer principle.
We have conducted comprehensive experiments using various real-world datasets and practical settings. GraphDIVE consistently advances the performance over various baselines and imbalanced learning methods. Further studies exhibited the rationality and effectiveness of GraphDIVE.




\nocite{langley00}

\bibliography{main}
\bibliographystyle{icml2021}





\clearpage
\appendix

\appendix
\onecolumn
\section{Influence of Expert Number}
\begin{figure}
		\centering
		\begin{subfigure}[t]{0.48\linewidth}
			\centering
			\includegraphics[width=\linewidth]{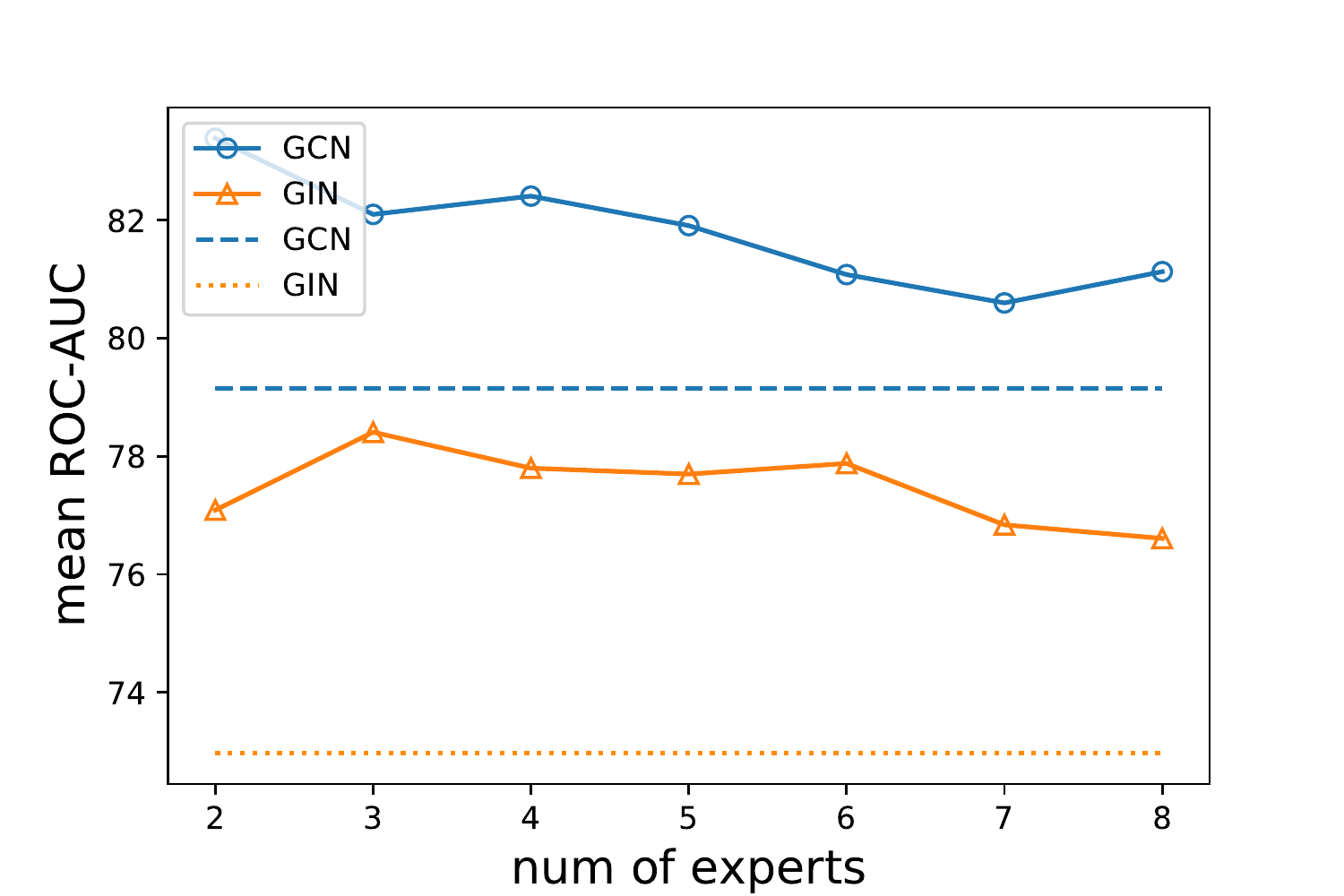} %
			\caption{\small \textbf{BACE}}
			\label{fig:bace}
		\end{subfigure}
		\hfill
		\begin{subfigure}[t]{0.48\linewidth}
			\centering
			\includegraphics[width=\linewidth]{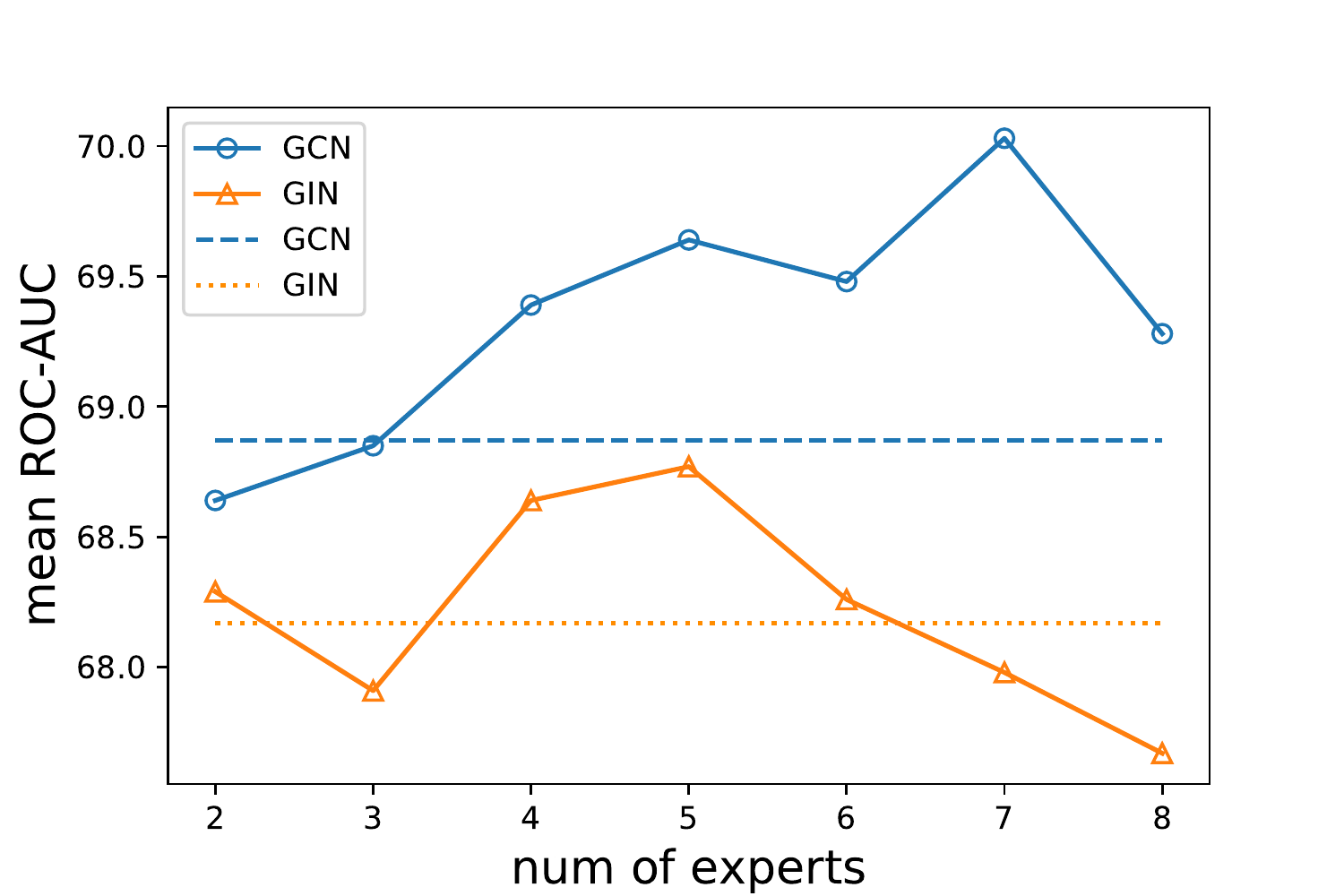} %
			\caption{\small \textbf{BBBP}}
			\label{fig:bbbp} 
		\end{subfigure}
		\vskip\baselineskip
		\begin{subfigure}[t]{0.48\linewidth}
			\centering
			\includegraphics[width=\linewidth]{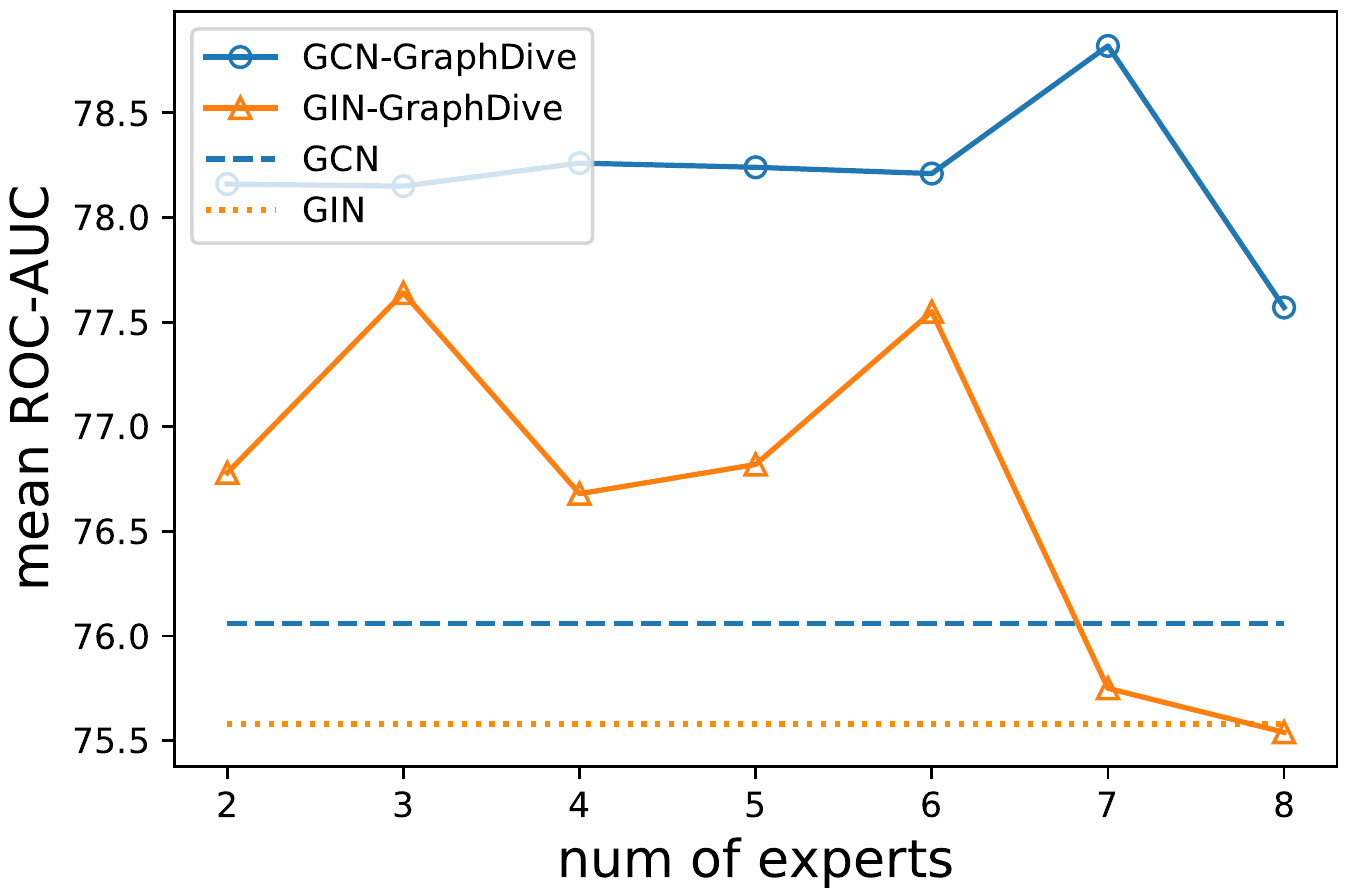} %
			\caption{\small \textbf{HIV}}
			\label{fig:hiv} 
		\end{subfigure}
		\hfill
		\begin{subfigure}[t]{0.48\linewidth}
			\centering
			\includegraphics[width=\linewidth]{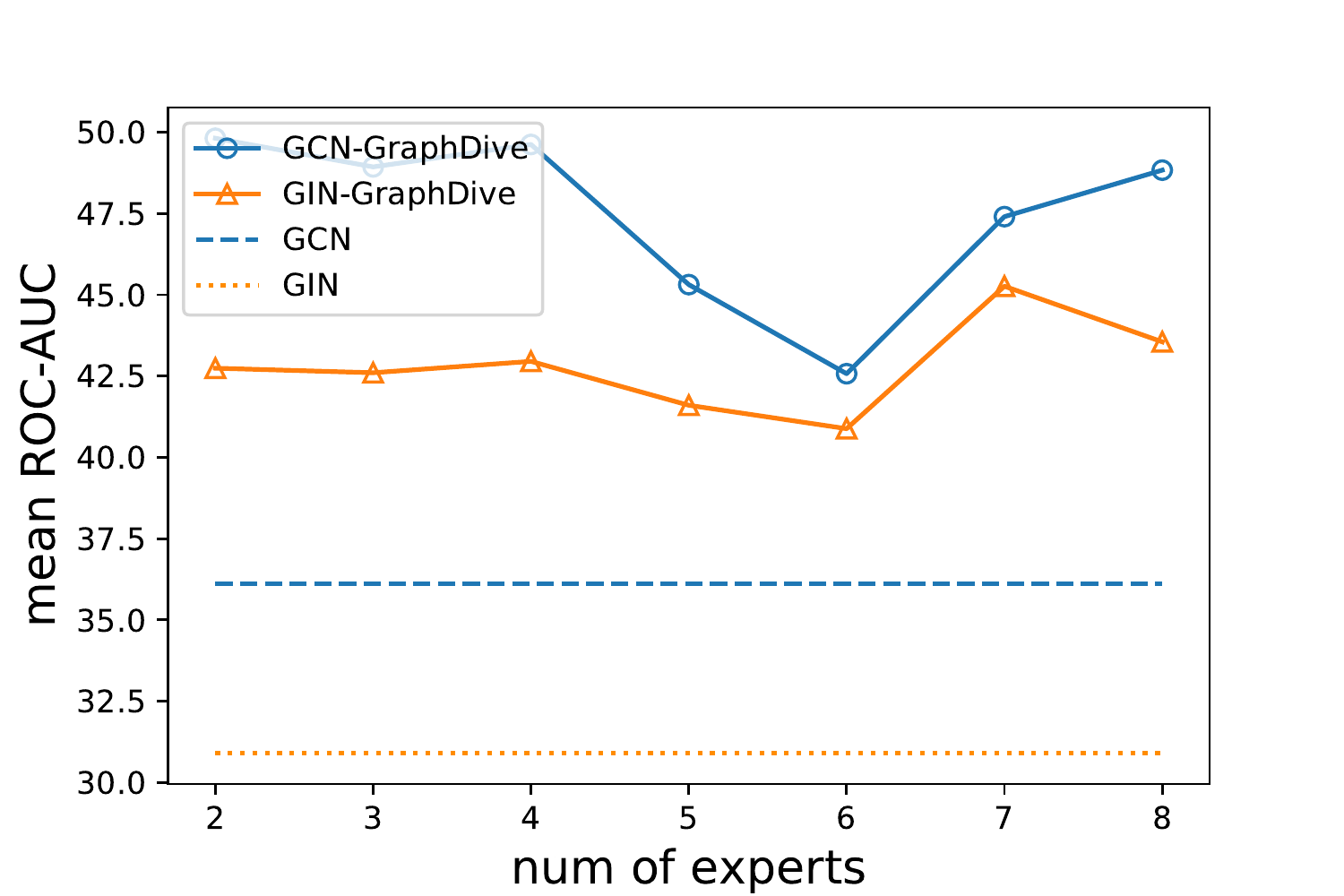} %
			\caption{\small \textbf{SIDER-3}}
			\label{fig:sider3}
		\end{subfigure}

		\caption{Mean ROC-AUC vs number of experts on four datasets: (a)BACE, (b)BBBP, (c)HIV, (d)SIDER-3}
	\label{fig:sensi}
	\end{figure}

We investigate the impact of expert number on GraphDIVE. To be more specific, we experiment with two to eight experts. The results on single-task datasets are shown in Figure \ref{fig:sensi}. From the figure, it can be found that GraphDIVE outperforms baseline methods at most of the  time, meaning that the expert numbers can be selected in a wide range. Besides, the performance improves with the increased number of experts at first, which demonstrates that more experts increase the model capacity. With the number of experts increasing, the experts which dominate the classification of minority class are more likely to generate unbiased predictions. For example, when the number of experts are less than eight, the performance of GraphDIVE-GCN  increases monotonously with the number of experts increasing on BBBP dataset.
Nevertheless, too many experts will inevitably introduce redundant parameters to the model, leading to over-fitting as well.

\section{Complexity Comparison}
In Table \ref{tbl:test_time}, we report the clock time of different methods on test set. For GraphDIVE, we employ GCN as a feature extractor and assign three experts on all datasets. It can be seen that GraphDIVE only introduces marginal computation complexity compared with GCN.


\begin{table}[h]
\centering
\caption{Complexity comparison between GraphDIVE and other GNNs. Test time (millisecond) are reported.}
\begin{tabular}{lcccc}
\toprule
          & BACE   & BBBP   & HIV      & SIDER-3 \\
\midrule
GCN       & 55.16  & 89.96  & 1248.01  & 54.98   \\
GIN       & 41.81  & 59.01  & 1019.72  & 40.95   \\
GCN+FLAG  & 86.76   & 90.22  & 1427.95  & 65.67   \\
GIN+FLAG  & 60.56  & 83.07  & 1224.01  & 60.63   \\
GSN       & 621.62 & 824.41 & 16100.56 & 640.24  \\
\midrule
GraphDIVE & 61.12  & 93.92  & 1267.65  & 64.78  \\
\bottomrule
\end{tabular}
\label{tbl:test_time}
\end{table}

\section{Experiments on multi-class Text Classification}
\begin{figure*}
    \centering
	\begin{subfigure}[t]{.32\linewidth}
			\centering
			\includegraphics[width=\linewidth]{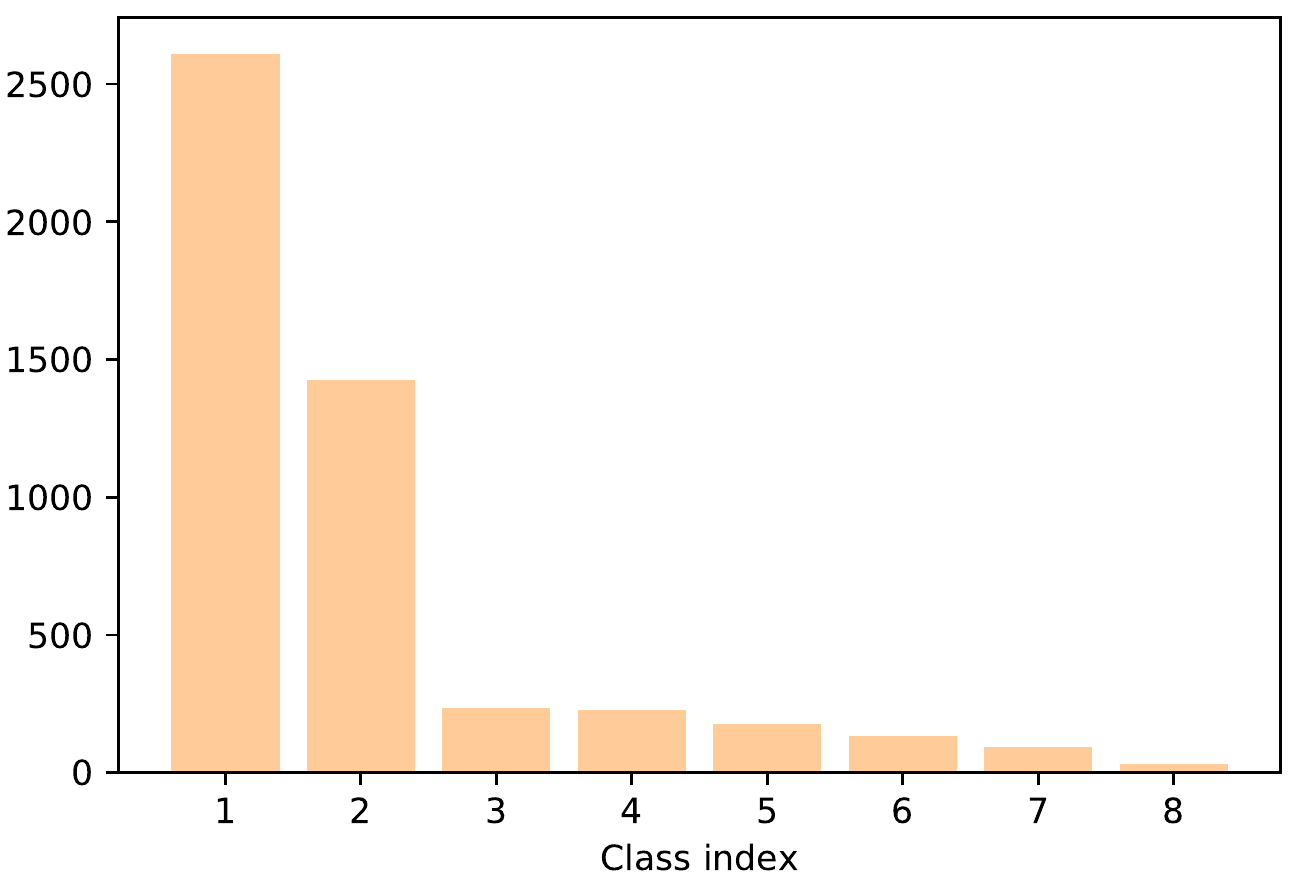} %
			\caption{\small \textbf{R8}}
	\label{fig:r8}
	\end{subfigure}
		\hfill
	\begin{subfigure}[t]{.32\linewidth}
			\centering
			\includegraphics[width=\linewidth]{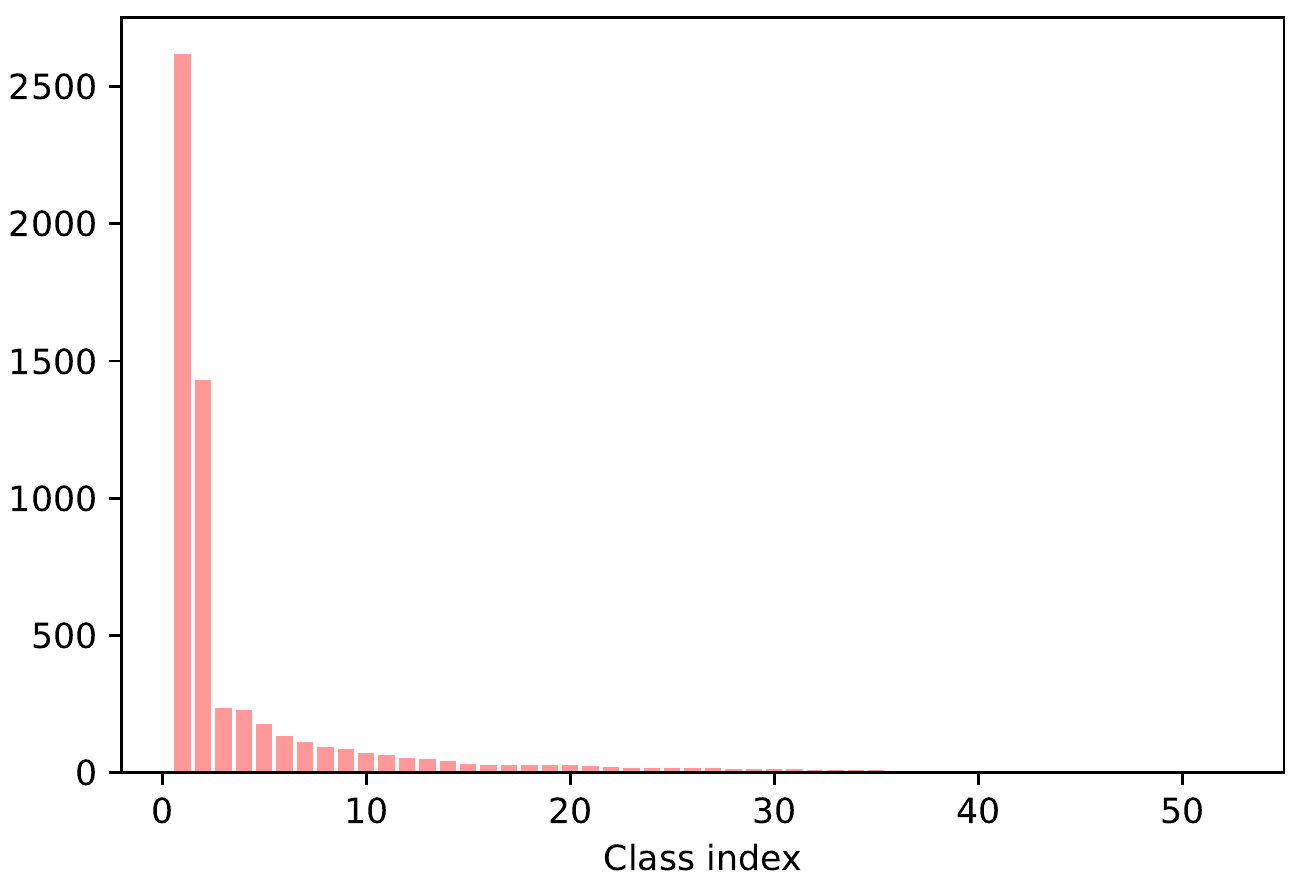} %
			\caption{\small \textbf{R52}}
			\label{fig:r52}
	\end{subfigure}
	\hfill
	\begin{subfigure}[t]{.32\linewidth}
			\centering
			\includegraphics[width=\linewidth]{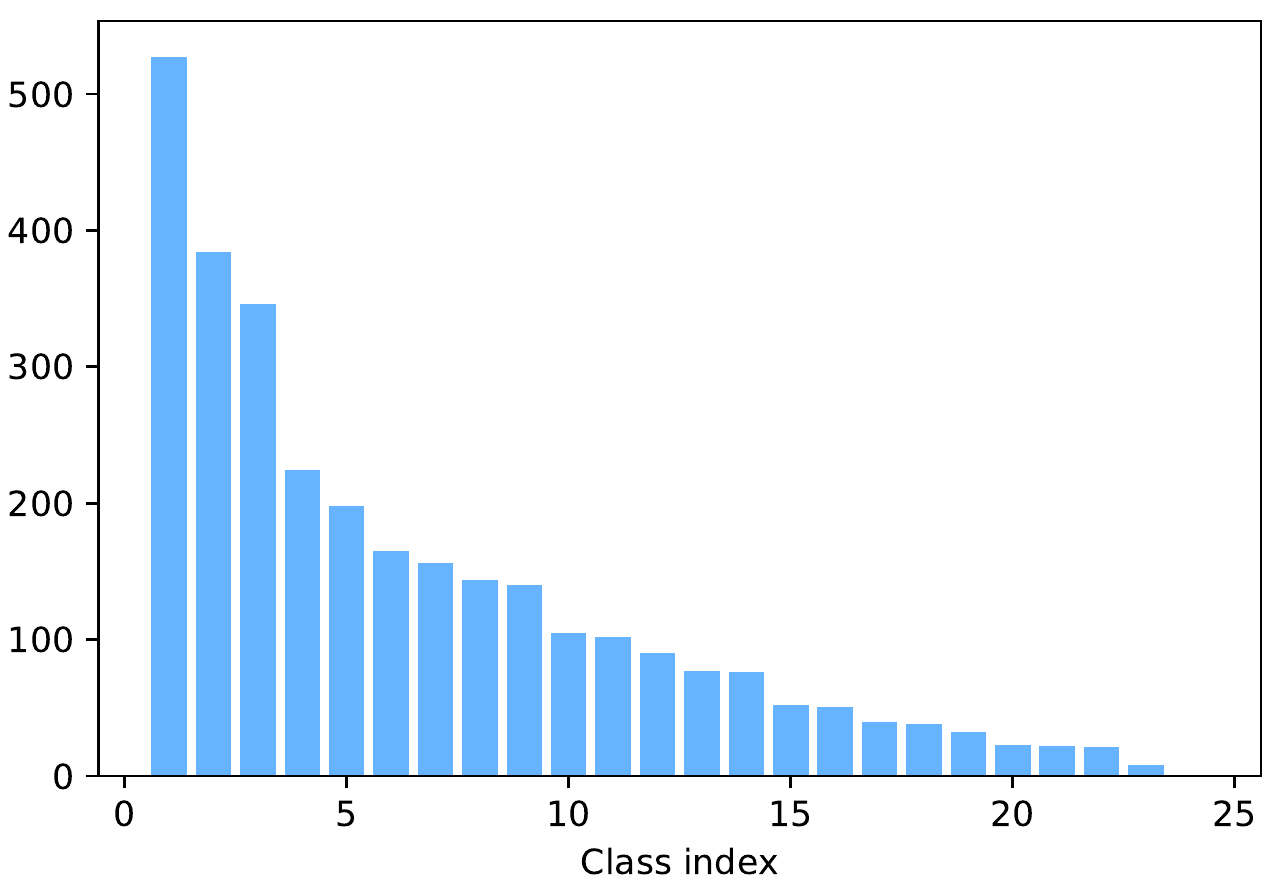} %
			\caption{\small \textbf{Oh}}
			\label{fig:oh}
	\end{subfigure}
	\hfill
\caption{An illustration of histograms on training class distributions for the datasets used in text-level graph classification experiment.}
\label{fig:text-dist}
\end{figure*}

In order to further verify that GraphDIVE is general for other imbalanced graph classification applications, we conduct experiments on text classification, which is an imbalanced multi-class graph classification task. Figure \ref{fig:text-dist} shows the class distribution of the widely used text classification datasets. It can be seen that these datasets have a long-tailed label distribution. \citet{yao2019graph} create a corpus-level graph which treats both documents and words as nodes in a graph. They calculate point-wise mutual information as word-word edge weights and use normalized TF-IDF scores as word-document edge weights. TextLevelGCN \cite{huang2019text} produces a text level graph for each input text, and transforms  text classification into graph classification task. TextLevelGCN fulfils the inductive learning of new words and achieves state-of-the-art results.

\begin{table}[h]
    \centering
    \footnotesize
    \caption{\label{acc-table} Accuracy on text classification datasets.  We run all models 10 times and report mean results. }
    \scalebox{1}{
    \begin{tabular}{cccc}
        \toprule
        \textbf{Model} & \textbf{R8}              & \textbf{R52}            & \textbf{Ohsumed}        \\ \midrule

        CNN           & 95.7 $\pm$ {0.5}          & 87.6 $\pm$ {0.5}          & 58.4 $\pm$ {1.0}          \\
        LSTM          & 96.1 $\pm$ {0.2}          & 90.5 $\pm$ {0.8}          & 51.1 $\pm$ {1.5}          \\
        Bi-LSTM       & 96.3 $\pm$ {0.3}          & 90.5 $\pm$ {0.9}          & 49.3 $\pm$ {1.0}          \\
        fastText      & 96.1 $\pm$ {0.2}          & 92.8 $\pm$ {0.1}          & 57.7 $\pm$ {0.5}          \\
        Text-GCN      & 97.0 $\pm$ {0.1}          & 93.7 $\pm$ {0.1}          & 67.7 $\pm$ {0.3}          \\
        TextLevelGCN     & {97.67} $\pm$ {0.2} & {94.05} $\pm$ {0.4} & {68.59} $\pm$ {0.7}       \\ 
        GraphDIVE     & \textbf{97.91} $\pm$ \textbf{0.2} & \textbf{94.31} $\pm$ \textbf{0.3} & \textbf{69.27} $\pm$ \textbf{0.6} \\ \bottomrule
    \end{tabular}
    }
    \label{tbl:text}
\end{table}

Hence, we use TextLevelGCN\footnote{\url{https://github.com/mojave-pku/TextLevelGCN}} \cite{huang2019text} as a feature extractor and closely follow the experimental settings as \citet{huang2019text}. From Table \ref{tbl:text}, we can see that GraphDIVE achieves better results across different datasets. This result indicates that GraphDIVE is not only applicable to molecular graphs, but also is beneficial to the imbalanced multi-class graph classification in text domain.

\end{document}